\documentclass[11pt]{article}

\usepackage[final]{acl}

\usepackage{times}
\usepackage{latexsym}

\usepackage[T1]{fontenc}
\usepackage{multirow}
\usepackage{amsmath}
\usepackage[utf8]{inputenc}
\usepackage{iftex}
\ifXeTeX
  \usepackage{xeCJK}
\fi
\usepackage{booktabs}
\usepackage{multirow}
\usepackage{array}
\usepackage{microtype}

\usepackage{inconsolata}

\usepackage{graphicx}

\title{Me-Agent: A Personalized Mobile Agent with Two-Level User Habit Learning for Enhanced Interaction}

\usepackage{hyperref} 
\author{
  Shuoxin Wang$^1$\footnotemark[1], ChangLiu$^2$\footnotemark[1],
  \textbf{Gowen Loo}$^3$\footnotemark[1], \textbf{Lifan Zheng}$^4$ \\
  \textbf{Kaiwen Wei}$^5$, \textbf{Yinyi Zeng}$^6$,
  \textbf{Jingyuan Zhang}$^7$, \textbf{Yu Tian}$^6$\footnotemark[2] \\
  $^1$Yunnan University \quad $^2$Hong Kong Polytechnic University \quad $^3$University of Electronic Science and Technology of China \\
  $^4$Southeast University \quad $^5$Chongqing University \quad
  $^6$Tsinghua University \quad $^7$Kuaishou Technology \\
  \texttt{shuoxin\_wang@163.com, tianyu1810613@gmail.com}
}

\begin{document}
\maketitle
{
  \renewcommand{\thefootnote}{\fnsymbol{footnote}} 
  
  \footnotetext[1]{Equal contribution.} 

  \footnotetext[2]{Corresponding author} 
}
\begin{abstract}

Large Language Model (LLM)-based mobile agents have made significant performance advancements. However, these agents often follow explicit user instructions while overlooking personalized needs, leading to significant limitations for real users, particularly without personalized context: (1) inability to interpret ambiguous instructions, (2) lack of learning from user interaction history, and (3) failure to handle personalized instructions. To alleviate the above challenges, we propose Me-Agent, a learnable and memorable personalized mobile agent. Specifically, Me-Agent incorporates a two-level user habit learning approach. At the prompt level, we design a user preference learning strategy enhanced with a Personal Reward Model to improve personalization performance. At the memory level, we design a Hierarchical Preference Memory, which stores users' long-term memory and app-specific memory in different level memory. To validate the personalization capabilities of mobile agents, we introduce User FingerTip, a new benchmark featuring numerous ambiguous instructions for daily life. Extensive experiments on User FingerTip and general benchmarks demonstrate that Me-Agent achieves state-of-the-art performance in personalization while maintaining competitive instruction execution performance. Our code is publicly available at: \url{https://github.com/Gianthaha/personal-mobile-agent.git}.

\end{abstract}

\section{Introduction}
\begin{figure}[htbp]
    \centering
    \includegraphics[
        width=1\columnwidth,
        keepaspectratio
    ]{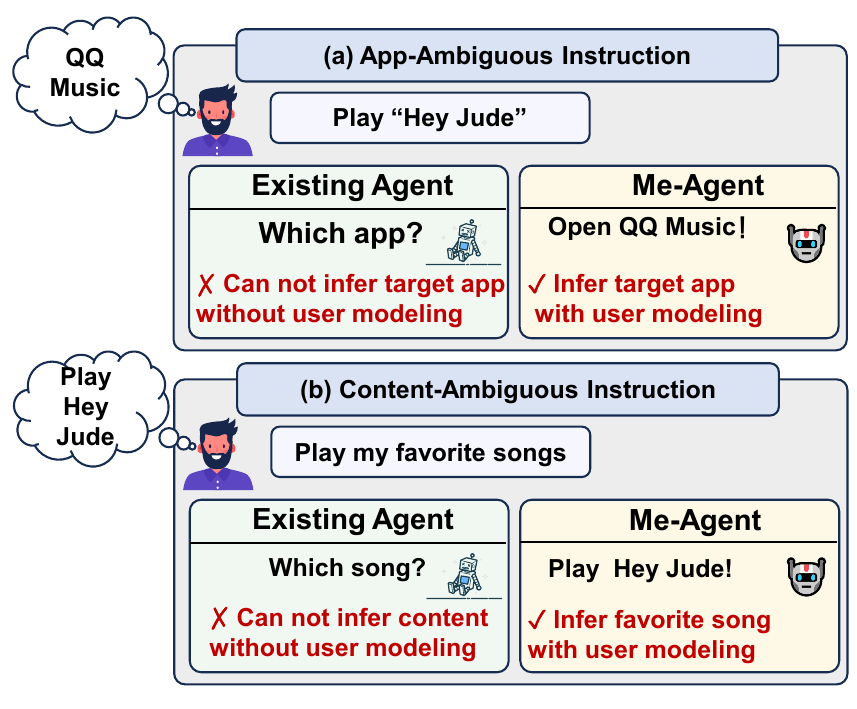}
    \caption{Two key challenges for mobile agents in ambiguous instruction.}
    \label{fig:question}
\end{figure}

With the continuous advancement of large language models (LLMs)~\cite{achiam2023gpt,anthropic_claude35_2024}, LLM-based mobile agent~\cite{qin2025ui-TARS,ye2025mobile-agent-v3} have achieved significant progress in performance and generalization, enabling them to execute complex tasks across various application scenarios. However, their learning paradigm still primarily relies on explicit user instructions, lacking systematic modeling and long-term adaptation to users' latent preferences and personalized needs. Particularly when user background information is absent or application scenarios are complex and dynamic, existing LLM-based mobile agents~\cite{wang2024mobile_v2,mobile_agent_e,li2024appagent_v2} exhibit three prominent limitations: 1) difficulty accurately interpreting natural language instructions with vague or implicit user intentions; 2) inability to continuously learn from and update user behavioral patterns across multi-turn interactions; and 3) limited capability in processing personalized instructions and preference configurations, hindering user-centric interactive experiences.

While prior research has made substantial progress in LLM personalization, applying these methods to mobile agents reveals significant limitations. First, existing methods~\cite{li2024personalized,zhang2025persona} often rely on fine-tuning, which proves impractical for mobile agents due to limited computational resources and storage. Cloud-based training alternatives introduce additional costs and raise privacy and security concerns. Second, many approaches~\cite{yao2025dprf,kong2024better} express user preferences and context through continuously expanding prompts, but without filtering mechanisms, this degrades inference efficiency and complicates storage and version management as interactions accumulate.

To address these limitations, we propose \textbf{Me-Agent}, a learnable and memorable personalized mobile agent that adapts to user preferences through two-level habit modeling without additional training. Me-Agent operates at two levels: 1) In prompt level, we introduce User Preference Learning strategy with a Personal Reward Model that ranks candidate responses by user preference alignment, guiding personalized decision-making. 2) In memory level, we construct a Hierarchical Preference Memory that stores and retrieves users' long-term memory and app usage patterns, mitigating prompt length issues while enabling continuous modeling of user trajectories and behavioral patterns. Me-Agent offers two key advantages: 1) it requires no model parameter modifications, making it suitable for API-based deployment; and 2) it employs a hierarchical optimization strategy that effectively mitigates memory decay.

To evaluate the personalization capabilities of mobile agents in executing ambiguous instructions, we construct User FingerTip to assess mobile agents' performance in inferring user intent, capturing personalized preferences, and adapting through multi-turn interactions. Extensive experiments demonstrate that Me-Agent achieves state-of-the-art performance on personalization metrics and surpasses baselines on performance metrics. We further evaluate Me-Agent on the E-dataset\cite{mobile_agent_e}, a general benchmark with complex, high-difficulty ,multi-app instructions. Me-Agent achieves a task completion rate of 89.3\%, a 39.7\% improvement over the baseline (63.9\%), demonstrating strong generalization to challenging mobile tasks. Our main contributions are as follows:
\begin{itemize}
\item {We design Me-Agent, a learnable and memorable personalized mobile agent that adapts to user preferences through two-level habit modeling without additional training.}
\item {We propose User Preference Learning, a parameter-free module for modeling user preferences by a Personal Reward Model.}
\item {We design Hierarchical Preference Memory, which stores users’ long-term and app-specific memory at different levels.} 
\item {We develop a personalized benchmark featuring two types of ambiguous instructions to test mobile agents' implicit preference reasoning.}
\end{itemize}

\section{Related Work}
\begin{table*}[t]

\centering
\setlength{\tabcolsep}{6pt}
\renewcommand{\arraystretch}{1.1}
\begin{tabular}{@{}llllcc@{}}
\toprule
\textbf{Type} & \textbf{Form} & \textbf{Instruction Example} & $app^{*}$ & $c^{*}$ \\
\midrule
Original
& $\langle app, a, c \rangle$
& ``Play  Hey Jude on QQ Music''
& -- & --  \\
Type~I
& $\langle \emptyset, a, c \rangle$
& ``Play Hey Jude''
 
& QQ Music
& -- \\
Type~II
& $\langle \emptyset, a, r \rangle$
& ``Play my favorite song''
 
& QQ Music
& Hey Jude \\
\bottomrule
\end{tabular}
\caption{Examples of instruction construction. $app^{*}$ and $c^{*}$ represent the ground-truth application and content.}
\label{tab:instruction_construction}
\end{table*}
\subsection{LLM-based Mobile Agents}

Early mobile agents use single-agent architectures~\cite{wang2024mobile_v1,zhang2025appagent}, which are prone to decision drift and error accumulation in long-horizon planning and cross-application tasks. Recent research on multi-agent architectures~\cite{wang2024mobile_v2,mobile_agent_e,li2024appagent_v2,loo2025mobilerag} focuses on improving instruction execution success rates through collaborative decision-making, hierarchical structures, self-evolution mechanisms, and structured knowledge bases for enhanced accuracy in complex scenarios. However, most existing mobile agents still rely on explicit user instructions and utilize limited user-related information, hindering their ability to predict user intent and model preferences. PerPilot~\cite{wang2025perpilot} introduces personalization by resolving user-specific entities via memory retrieval, but it primarily focuses on entity resolution and does not capture long-term user behavior or preferences. In contrast, Our work focuses on modeling user behavioral preferences in mobile agents to enable more personalized instruction execution.

\subsection{Personalization in LLMs} 

Personalization of large language models is important research direction, which can be categorized into four main approaches~\cite{zhang2024personalizationsurvey}.
Parameter-level methods ~\cite{hu2022lora,dettmers2023qlora} enable efficient personalization through fine-tuning techniques, while prompt and context-driven methods~\cite{salemi2024lamp,zhuang2024hydra} use user information as prompts to guide model outputs. Additionally, memory and retrieval-augmented methods~\cite{fan2025towards,CFRAG,maragheh2025arag} enhance personalization by maintaining long-term user memory, and preference alignment methods~\cite{ouyang2022RLHF,li2024P-RLHF,poddar2024personalizing} leverage reinforcement learning to model diverse user preferences effectively. While these methods effectively mitigate the personalization issues of LLMs, their application in the LLM-based agent encounters challenges like training difficulties and contextual redundancy. To overcome these challenges, we propose Me-Agent, a personalized mobile agent framework that models user habits at both the prompt and memory levels, enabling implicit intent inference from ambiguous instructions without training.

\section{User FingerTip}

Real-world users tend to use ambiguous expressions—omitting application names (e.g., "I want to listen to music") or using imprecise references (e.g., "play what I usually play"). Such instructions require the agent to infer application and content preferences from the user's behavioral history. However, existing personalized datasets~\cite{yang2025fingertip,wang2025perpilot} fail to assess this capability due to two major limitations: 1) personalized information is explicitly stated in the commands; 2) the small scale of the datasets limits comprehensive capability evaluation. 
To fill this gap, we extend \textbf{FingerTip} to build a new personalized benchmark, introducing two types of ambiguous instructions to test mobile agents' implicit preference reasoning.

\subsection{Data Construction}
The basic instruction can be formalized as a triplet\begin{equation}
    I_{orig} = \langle app, a, c \rangle,
\end{equation}
where
$app$ denotes the target application, $a$ denotes the action, and $c$ denotes the operation content. In real scenarios, users tend to use ambiguous expressions—omitting application names (e.g., "I want to listen to music") or replacing specific content with vague references (e.g., "order my usual"). As shown in Table~\ref{tab:instruction_construction}, we design two types of ambiguous instructions to construct User FingerTip. 

\noindent\textbf{Type I: App-Ambiguous Instruction.}
Application preference reasoning is a fundamental capability for personalized mobile agents, as multiple applications often coexist within the same functional category (e.g., NetEase Music and QQ Music). However, existing datasets~\cite{mobile_agent_e,zhang2025appagent,wang2024mobile_v2} explicitly provide target applications in instructions, making it infeasible to evaluate whether mobile agents can infer user preferences from behavioral history. To address this, we construct Type~I instructions by removing the application name:
\begin{equation}
I_{\text{type I}} = \langle \emptyset, a, c \rangle, 
\end{equation}
where $a$ denotes the user action and $c$ denotes the explicit content.   The ground truth is annotated as the user's most frequently used application within the corresponding category.

\begin{table}[t]
\centering
\setlength{\tabcolsep}{6pt}
\renewcommand{\arraystretch}{1.15}
\begin{tabular}{@{}l c@{}}
\toprule
\textbf{Statistic} & \textbf{Value} \\
\midrule
\ Users & 60 \\
\ Functional Categories & 12 \\
\ Applications & 33 \\
\ Type I Instructions & 300  \\
\ Type II Instructions & 267 \\

\ Training Instructions per User & 15 \\
\ Test Instructions per User & 5 \\
\bottomrule
\end{tabular}
\caption{Dataset statistics.}
\label{tab:statistics}
\end{table}

 \begin{figure*}[t]
    \centering
    \includegraphics[width=\textwidth]{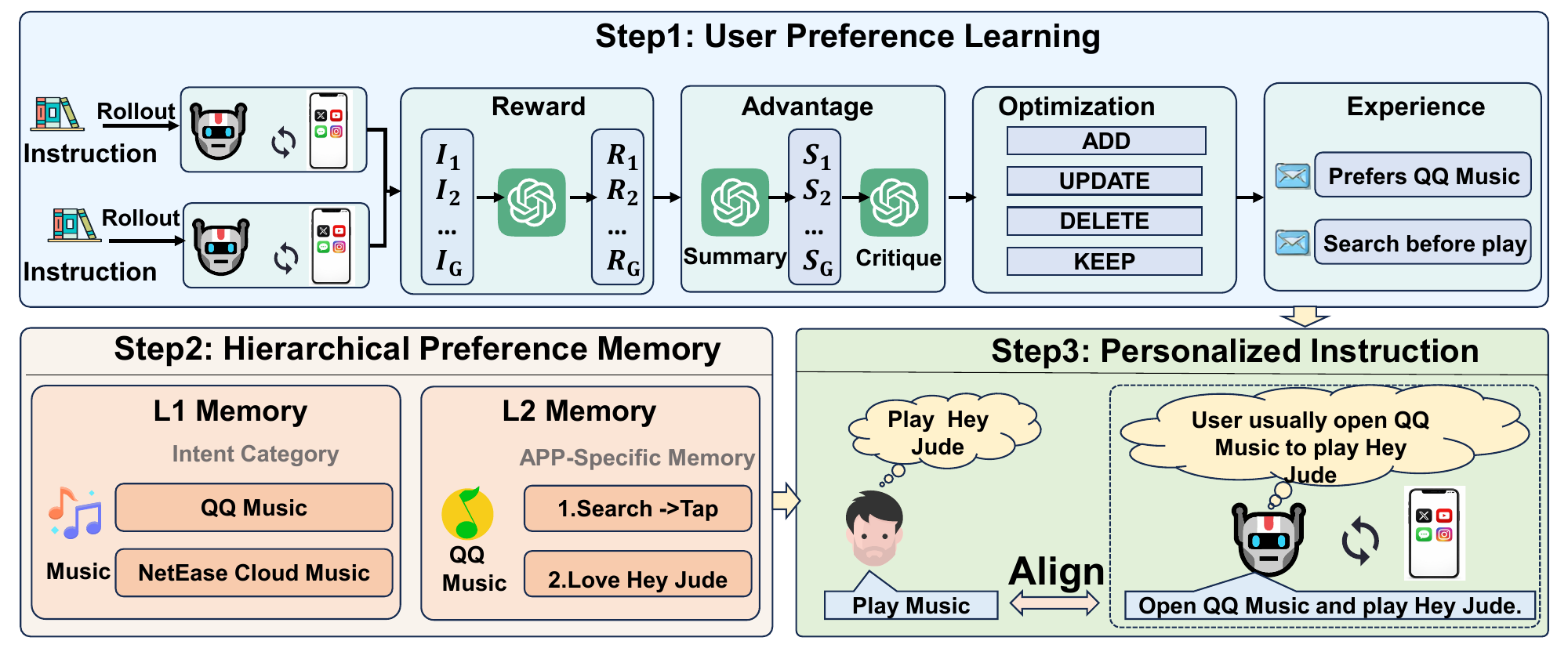}
    \caption{Overview of the proposed Me Agent framework.}
    \label{fig:method}
\end{figure*}
\noindent\textbf{Type II: Content-Ambiguous Instruction.}
While Type~I focuses  on application preference reasoning, comprehensive personalization also requires understanding user content preferences. We design Type~II instructions to evaluate joint reasoning: inferring the target application while retrieving preferred content from behavioral history. Specifically, we extend Type~I by replacing explicit content $c$ with implicit references $r$ (e.g., ``my favorite'', ``the usual''), yielding:
\begin{equation}
    I_{\text{type II}} = \langle \emptyset, a, r \rangle.
\end{equation}
Each instruction is annotated with both the target application $a^{*}$ and the user-preferred content $c^{*}$ inferred from historical behavior.

\subsection{Data Statistics}

Table~\ref{tab:statistics} summarizes the dataset statistics. User FingerTip includes 60 users, 33 applications across 12  categories, with 300 Type~I and 267 Type~II instructions in total. We partition the dataset at the user level: for each user, 15 instructions are randomly sampled for training and 5 for testing. The training set enables agents to learn user preferences, while the test set evaluates preference reasoning capabilities. Training and test instructions are non-overlapping to ensure evaluation validity.

\section{Methodology}  
\subsection{Problem Formulation}

Given user instruction $I$, user preference profile $U$,  historical action sequence $A_{\text{history}} = \{a_1, \ldots, a_t\}$,  current screenshot observation $O_t$, and  available action space $\mathcal{A}_t$, the mobile agent predicts the next action as:
\begin{equation}
a_{t+1} = f(I, A_{\text{history}}, O_t, \mathcal{A}_t , U),
\end{equation}
where $f(\cdot)$ denotes the agent. It iterates this process until instruction $I$ is completed.

\subsection{Me-Agent}

Current LLM-based agents~\cite{mobile_agent_e,maragheh2025arag} face the following challenges: 1) a lack of capability for modeling user preferences, making it difficult to understand users' vague intent in instructions; 2) limitations in mobile training, as the computational power of mobile devices is insufficient to support local training; and 3) difficulties in tracking the dynamic evolution of preferences, due to a lack of cross-iteration continuous learning mechanisms, which hampers the ability to update user behavior dynamically, resulting in a decline in personalization effectiveness over time. To address the above issues, we propose Me-Agent, a training-free personalized mobile agent framework. As shown in the Figure~\ref{fig:method}, Me-Agent is composed  of \textbf{User Preference Learning} and \textbf{Hierarchical Preference Memory}.

\subsection{User Preference Learning}
Inspired by Training-Free GRPO~\cite{cai2025trainingfreegrpo}, we propose User Preference Learning (UPL), a parameter-free module for modeling user preferences. Specifically, UPL includes two targeted improvements: (1) a VLM-based reward model that evaluates task completion from visual observation, and (2) predefined experience categories for structured preference accumulation. Built upon these designs, UPL operates through four stages: Rollout, Reward, Advantage, and Optimization.

\noindent\textbf{Rollout.} For each user instruction $I$, we perform $G$ independent rollouts using the mobile agent $f(\cdot)$ to obtain diverse execution trajectories that provide comparative samples for subsequent intra-group analysis:
\begin{equation}
\{\tau_1, \tau_2, ..., \tau_G\} \sim f(I, \mathcal{E}),
\end{equation}
where $\mathcal{E}$ denotes the current experience pool. Each trajectory $\tau_g = \{(O_t^g, A_t^g)\}_{t=1}^{T_g}$ represents a complete instruction execution, comprising a sequence of screenshots and corresponding actions. A sampling temperature $T > 0$ is applied to introduce exploration diversity, ensuring coverage of different operational paths.

\noindent\textbf{Reward.} We design a VLM-based reward model that leverages a Vision-Language Model to assess instruction execution from screenshot sequences. The reward model offers several advantages: 1) It is easily accessible, extracting information directly from screenshots, facilitating data collection across various application scenarios. 2) It provides enriched feedback, reflecting instruction execution completeness levels and catering to personalized information. It is defined as:
\begin{equation}
    r_g = R_{\text{VLM}}(I, \{O_1^g, O_2^g, \ldots, O_{T_g}^g\}),
\end{equation}
where $I$ denotes the instruction and $\{O_1^g, \ldots, O_{T_g}^g\}$ denotes the screenshot sequence. The VLM evaluates four dimensions: goal achievement, step validity, result visibility, and error detection. The reward $r_g \in [0, 1]$ provides a continuous signal for fine-grained trajectory quality assessment.

\noindent\textbf{Advantage.}
The personalized mobile agent needs not only to identify which trajectory is superior but also to understand user preference patterns. By having an LLM analyze successful and failed trajectories, we extract natural language experiences that are interpretable, editable, and can be directly injected into prompts to guide subsequent decisions. Given the trajectory set $\{(\tau_g, r_g)\}_{g=1}^{G}$ for instruction $I$, we extract experiences through two phases:

\textbf{Trajectory Summarization.} Each trajectory is summarized into structured experiences:
\begin{equation}
S_g = f( I, \tau_g, r_g).
\end{equation}

\textbf{Comparative Critique.} All operational summaries for the same instruction are compared to extract generalizable experiences:
\begin{equation}
\mathcal{E}_{\text{new}}^I =f(  I, \{S_g\}_{g=1}^{G}).
\end{equation}
The critique process focuses on six experience categories: user preferences, UI navigation, action sequences, element identification, context awareness, and completion signals. 

\noindent\textbf{Optimization.}
As learning iterates, newly extracted experiences may overlap or conflict with existing ones. Direct appending leads to redundant pool expansion, while naive overwriting risks losing valuable historical memory. To address this, we design an integration mechanism that enables agents to compare new experiences against the existing pool and assigns each one an operation:
\begin{equation}
\{op_j\} = f(  \mathcal{E}_{\text{old}}, \mathcal{E}_{\text{new}}),
\end{equation}
where $op_j \in \{\texttt{ADD}, \texttt{UPDATE}, \texttt{Delete}, \texttt{KEEP}\}$ denotes inserting new information, refining existing experiences,  deleting redundant ones, or remaining unchanged, respectively. 


\subsection{Hierarchical Preference Memory}
The User Preference Learning module produces an experience pool $\mathcal{E}$ that stores general user preferences and is directly injected into prompts. However, app-specific knowledge is extensive and only relevant when interacting with the corresponding app. Injecting all such memory into prompts would cause context overflow and attention dilution. To address this, we introduce Hierarchical Preference Memory (HPM), which stores application-specific memory externally and retrieves it as needed for each target application.
 
\noindent\textbf{Personalized Memory.}
Mobile instructions exhibit a natural hierarchical structure: users first select an app category (e.g., music, navigation), then perform app-specific operations. We design a hierarchical memory accordingly:

\textbf{Level-1 Memory $\mathcal{M}^{L1}$} is organized by intent categories:
\begin{equation}
\mathcal{M}^{L1} = \{(c_i,  , \mathcal{A}_i)\}_{i=1}^{C},
\end{equation}
where $c_i$ denotes the category nameand $\mathcal{A}_i$ is the set of apps under this category.

\textbf{Level-2 Memory $\mathcal{M}^{L2}$} stores app-specific memory for each app:
\begin{equation}
\mathcal{M}^{L2}_{app} = \{\text{workflows}, \text{preferences}\},
\end{equation}
including workflows (instruction execution patterns and UI element positions) and the content preferences (user-specific preferences within the app).

For successful trajectories with reward $r_g \geq \theta$, we use an LLM to extract workflows and content preferences from the execution trace. The memory supports dynamic updates: similar workflows are merged based on embedding similarity and success counts, while content preferences are adjusted using frequency statistics to reflect evolving user habits.

\noindent\textbf{Personalized Inference.}
During inference, the mobile agent must resolve two types of ambiguity in user instructions: \textit{application ambiguity} (which app to use) and \textit{content ambiguity} (what specific content to operate on). We design a two-stage inference mechanism that leverages the learned experience pool $\mathcal{E}$ and user memory $\mathcal{M}$.

\begin{table*}[t]
\centering
\begin{tabular}{clccccccc}
\toprule
\textbf{Type} &
\textbf{Methods} &
\textbf{ASA ↑} &
\textbf{BERTScore ↑} &
\textbf{PS ↑} &
\textbf{TCR ↑} &
\textbf{TSR ↑} &
\textbf{RF ↑} &
\textbf{AF ↑} \\
\midrule
\multirow{3}{*}{Type I} 
 & Mobiel Agent V2           & 0.600 & — & — & 0.369 & 0.200 & 0.464 & 0.698 \\
 & Mobiel Agent E  & 0.400 & — & — & 0.763 & 0.600 & \textbf{0.917} & \textbf{0.950} \\
 & Me-Agent     &  \textbf{1.000} & — & — & \textbf{0.960} & \textbf{0.800}  & 0.870 & 0.855 \\
\midrule
\multirow{3}{*}{Type II} 
 &  Mobiel Agent V2  &0.200 & 0.526  & 0.265 & 0.778 & 0.300 & 0.838 & 0.898 \\
 & Mobiel Agent E & 0.200  & 0.549 & 0.275 & 0.543 & 0.125 & 0.838 & \textbf{0.900} \\
 & Me-Agent    &\textbf{1.000} & 0.725 & 0.534 & \textbf{0.888} & \textbf{0.500} & \textbf{0.935} & 0.876 \\
\bottomrule
\end{tabular}
\caption{Personalized evaluation on the User FingerTip.}
\label{tab:overall_results}
\end{table*}

\paragraph{Application Resolution.} 
When instruction $I$ explicitly specifies an application, the agent uses it directly. Otherwise, it infers the target application based on user preferences:
\begin{equation}
a^* =
\left\{
\begin{array}{ll}
a, & \text{if } a \in I, \\[4pt]
\mathop{\arg\max}\limits_{a \in \mathcal{A}_{\text{cat}(I)}} P(a \mid \mathcal{E}), & \text{otherwise}.
\end{array}
\right.
\end{equation}
where $\mathcal{A}_{\text{cat}(I)}$ denotes the set of applications within the functional category of instruction $I$, and $P(a \mid \mathcal{E})$ represents the user's preference probability for application $a$ derived from the experience pool.

\textbf{Content Retrieval.} When the instruction contains implicit references (e.g., ``my favorite'',) the mobile agent retrieves relevant content from HPM through a two-step process. First, we retrieve the top-$k$ candidate contents $\mathcal{C}_{\text{cand}}$ based on semantic similarity:
\begin{equation}
    \mathcal{C}_{\text{cand}} = \text{Top}_k \left( \text{sim}(\mathbf{e}_I, \mathbf{e}_c) \mid c \in \mathcal{M}_{a^*} \right),
\end{equation}
where $\mathcal{M}_{a^*}$ denotes the memory associated with the target application and $\text{sim}(\cdot, \cdot)$ measures the semantic similarity between embeddings. Then, the LLM selects the most appropriate content by reasoning over the candidates and user history:
\begin{equation}
    c^* =f(I, \mathcal{C}_{\text{cand}}).
\end{equation}

\begin{table}[t]
\resizebox{\linewidth}{!}{
\centering
\begin{tabular}{l|cccc}
\toprule
\textbf{Method} & \textbf{TCR ↑} & \textbf{TSR ↑} & \textbf{AF ↑} & \textbf{RP ↑} \\
\midrule
Mobile-Agent-v2 & 0.414 & 0.200 & 0.467 & 0.570 \\
Mobile-Agent-E & 0.639 & \textbf{0.667} & 0.617 & 0.702 \\
Me Agent & \textbf{0.893} & 0.600 & \textbf{0.861} & \textbf{0.978} \\
\bottomrule
\end{tabular}}
\caption{Performance Evaluation on E Dataset.}
\raggedright
\footnotesize
\label{tab:results}
\end{table}

\textbf{Prompt Injection.} The resolved application $a^*$, retrieved content $c^*$, and relevant experiences from $\mathcal{E}$ are injected into the prompt to enable personalized execution based on user preferences.

\begin{table*}[t]
\centering
\begin{tabular}{clccccccc}
\toprule
\textbf{Type } &
\textbf{Backbone} &
\textbf{ASA ↑} &
\textbf{BERTScore ↑} &
\textbf{PS ↑} &
\textbf{TCR ↑} &
\textbf{TSR ↑} &
\textbf{RF ↑} &
\textbf{AF ↑} \\
\midrule
\multirow{3}{*}{Type I} 
 & GPT-4o            & 1.00 & — & — & 0.976 & 0.833 & 0.933 & 1.000 \\
 & Claude-3.5-Sonnet & 1.00 & — & — & 0.920 & 1.000 & 0.935 & 0.975 \\
 & Gemini-2.5-Pro    & 1.00 & — & —  &1.00 & 0.958 & 1.000 & 1.000 \\
\midrule
\multirow{3}{*}{Type II} 
 & GPT-4o            & 1.00 & 0.554 & 0.800 & 0.900 & 0.600 & 0.825 & 0.825 \\
 & Claude-3.5-Sonnet & 1.00 & 0.534 &  0.850 & 0.697 & 0.400 & 0.733 & 0.783 \\
 & Gemini-2.5-Pro    & 1.00 & 0.534 &  0.750 & 0.893 & 0.937 & 0.933 & 0.893 \\
\bottomrule
\end{tabular}
\caption{Evaluation under different reasoning backbones.}
\label{tab:llm_results}
\end{table*}

\section{Experiments} 
\subsection{Experimental Setup}
\noindent\textbf{Dataset \& Metrics.} We assess personalization capabilities on User FingerTip and general capabilities on E Datasets \cite{mobile_agent_e}. We evaluate agents from two perspectives. \textbf{Personalization metrics} include: \textbf{App Selection Accuracy (ASA)}, measuring whether the agent selects user-preferred apps among functionally equivalent options; \textbf{BERTScore}, computing semantic similarity between agent outputs and ground-truth references (Type~II only); and \textbf{Preference Score (PS)}, evaluating reasoning alignment with user preferences via LLM scoring (Type~II only). \textbf{Performance metrics} comprise: \textbf{Task Completion Ratio (TCR)}, measuring the proportion of achieved sub-goals; \textbf{Task Success Rate (TSR)}, a binary metric requiring all requirements to be fulfilled; \textbf{Action Fidelity (AF)}, assessing correctly executed atomic actions; and \textbf{Reflection Precision (RP)}, quantifying the rate of correct diagnoses during error recovery.

\noindent\textbf{Baseline.} We compare our approach against two baselines: Mobile-Agent-v2 ~\cite{wang2024mobile_v2} and Mobile-Agent-E~\cite{mobile_agent_e}. To ensure a fair comparison, we evaluate Me-Agent against the baselines using the same atomic action space and perception models. We use DBNet~\cite{DBNET} and the ModelScope ConvNextViT-document for OCR detection and recognition. We utilize GroundingDINO~\cite{liu2024grounding} for icon grounding and GPT-4o~\cite{hurst2024gpt} for generating captions for each cropped icon.

\noindent\textbf{Implementation Details.} We conduct experiments on standardized cloud infrastructure. We use a group size of 2 and sampling temperature of 0.3. Models are trained for 2 epochs with batch size 5. We ensure reproducibility by maintaining identical software versions and hardware configurations across all experimental runs.

\subsection{Personalized Evaluation}

We conduct a comprehensive evaluation of Me-Agent on the User FingerTip. As shown in Table~\ref{tab:overall_results}, It significantly outperforms Mobile-Agent-v2 and Mobile-Agent-E on personalization metrics, while also demonstrating superior performance on task execution metrics.

\noindent\textbf{Personalization Capabilities.}
Equipped with UPL and HPM, Me-Agent learns  preferences from user historical behavior, achieving an ASA of 1.000 on both Type I and Type II instructions. In contrast, Mobile-Agent-v2 and Mobile Agent E lack preference modeling capabilities, which likely contributes to their lower ASA when facing multiple applications within the same functional category (e.g., QQ Music vs. NetEase Music). Furthermore, Me-Agent achieves notably higher BERTScore and PS on Type II instructions, indicating its ability to infer user content preferences. These results confirm the learning and memory capabilities of the Me-Agent in user personalization: it captures application-level preferences through an experience pool, while hierarchical memory enables personalization at the content level.

\noindent\textbf{Instruction Execution Performance.}  Me-Agent achieves the highest TCR and TSR across both instruction types. Notably, as instruction complexity increases from Type I to Type II, the baselines show substantial performance degradation, while Me-Agent maintains stable performance. We attribute this robustness to the HPM module, which records UI element positions and action sequences from successful interactions, enabling the agent to reuse proven operational paths rather than exploring complex interfaces from scratch.

\subsection{Performance Evaluation}

We evaluate the performance of Me-Agent on the E dataset\cite{mobile_agent_e}. As shown in Table \ref{tab:results}, Me-Agent achieves the best performance on Task Completion Rate (0.893), Action Accuracy (0.861), and Reflection Precision (0.978). For Task Success Rate, Me-Agent obtains 0.600, which is slightly lower than Mobile-Agent-E (0.667) but substantially higher than Mobile-Agent-v2 (0.200).

We attribute these improvements to Me-Agent’s experience learning mechanism. Mobile-Agent-v2 and Mobile-Agent-E often fail not due to incorrect planning but because of inaccurate UI element recognition. When visual models misidentify UI element positions or semantics, correct plans cannot be reliably executed, a problem further exacerbated by frequent application updates.

\begin{figure}[t]
    \centering
    \includegraphics[width=1\columnwidth]{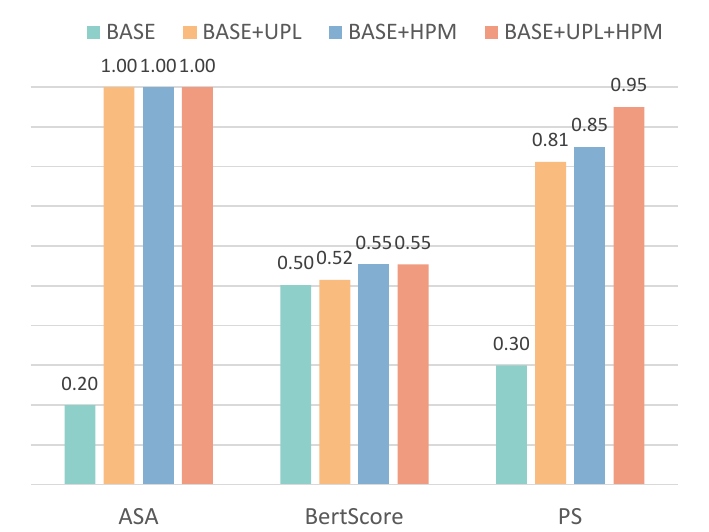}
    \caption{Ablation Study on the User Fingertip. }
    \label{fig:data}
\end{figure}

\begin{figure*}[t]
    \centering
    \includegraphics[width=1\textwidth]{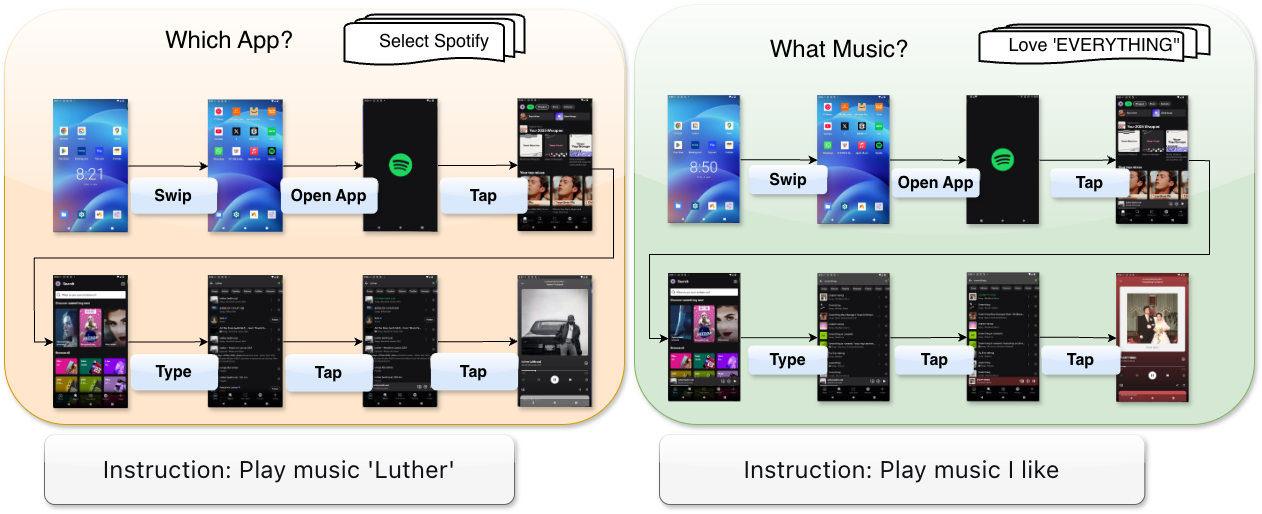}
    \caption{Example cases of Me-Agent executing personalized mobile instructions}
    \label{fig:case_study}
\end{figure*}

Me-Agent addresses this issue by learning from past executions. \textbf{UPL} explores different ways to complete the same task, while the Advantage stage compares successful and failed attempts to extract useful knowledge. This knowledge includes three key types: 1) action sequences that record correct operation orders, 2) UI navigation that stores element positions and paths, and 3) completion signals that define when a task is finished. \textbf{HPM} organizes this knowledge by application, allowing the agent to retrieve relevant experience during inference. With this experience, Me-Agent can locate UI elements more accurately and execute tasks more reliably.

\subsection{Ablation Study}
We conduct ablation studies to analyze the contributions of the proposed \textbf{UPL} and \textbf{HPM} modules. The base method is vanilla Mobile-Agent-E. We then incrementally incorporate the \textbf{UPL} module (\textbf{BASE+UPL}), the \textbf{HPM} module (\textbf{BASE+HPM}), and their combination (\textbf{BASE+UPL+HPM}). As shown in Figure~\ref{fig:data}: 1)Mobile-Agent-E exhibits clear limitations in personalized settings, validating our motivation that current agents lack the ability to model user preferences. 2) UPL leads to a substantial improvement in task completion, demonstrating the effectiveness of shifting policy optimization from parameter space to context space. 3)HPM significantly enhances semantic quality and preference alignment, indicating that hierarchical memory enables more accurate resolution of ambiguous instructions through on-demand retrieval of application-specific knowledge. 4)The two modules are complementary: UPL captures general user preferences, while HPM provides fine-grained execution knowledge. Their combination achieves the best overall performance.

\subsection{Impact of Backbone Models}
We evaluate Me-Agent using three backbone models: \textbf{GPT-4o}, \textbf{Claude-3.5-Sonnet}, and \textbf{Gemini-2.5-Pro}. As shown in Table \ref{tab:llm_results}, our method achieves stable personalized evaluations and task performance across all models for both Type I and Type II tasks. This robustness stems from the modular design of Me-Agent: the UPL module encodes preferences as natural language experiences that are compatible with any LLM, while the HPM module leverages structured retrieval to reduce dependence on backbone-specific capabilities.

\subsection{Case Study}

Figure \ref{fig:case_study} illustrates the personalization capabilities of Me-Agent in handling ambiguous instructions. When a user says, "Play the song 'Luther,'" Me-Agent directly identifies and plays the correct track from Spotify by analyzing historical listening habits, rather than requesting clarification as conventional mobile agents do~\cite{li2024appagent_v2,mobile_agent_e}. For vague instructions such as "Play the music I like," it automatically recommends content aligned with the user's preferences by synthesizing listening history and preferred genres, suggesting songs like "EVERYTHING," "Luther," and "Hey, Judy." This seamless interaction eliminates redundant confirmations and enhances the accuracy of processing ambiguous instructions.

\section{Conclusions}

To alleviate the challenges faced by LLM-based mobile agents in interpreting ambiguous instructions, learning from interaction history, and handling personalized tasks, we propose Me-Agent. It incorporates User Preference Learning, which enables preference modeling without parameter updates by shifting optimization to context space, and Hierarchical Preference Memory, which retrieves application-specific knowledge on demand to prevent context overload. To evaluate personalization capabilities, we introduce User FingerTip, a benchmark comprising ambiguous daily-life instructions. Extensive experiments demonstrate that Me-Agent outperforms the baselines in both personalization and task performance.

\newpage
\section*{Limitations}

\paragraph{UI Dynamics and Temporal Validity}. We extract UI positions and action sequence from successful task executions. However, these learned experiences may become invalid when the application is updated or unexpected pop-ups appear. Such UI changes can cause execution failures because the stored positions no longer match the current interface.

 \paragraph{Limited Contextual Factors in Personalization}. Our personalization approach mainly relies on users' historical preferences and behavior patterns. It does not consider dynamic factors such as the user's current location or emotional state. This may lead to poor recommendations when the user's current situation differs from their usual patterns.

\bibliography{custom}

\clearpage
\appendix 

\section{ Prompt }
\label{sec:appendix_prompts}
In this appendix, we provide the system and user prompts used in mobile UI automation tasks. 
Tables \ref{tab:rollout_prompt} and \ref{tab:single_query_critique_prompt} present the prompts for trajectory analysis and single-task experience extraction. 
Tables \ref{tab:group_experience_update_prompt} and \ref{tab:batch_experience_update_prompt} are used for experience integration and global experience consolidation. 
Table \ref{tab:problem_with_experience_prompt} works for guiding task execution based on learned experiences.

\begin{table*}[t]
\centering
\setlength{\tabcolsep}{4pt}
\renewcommand{\arraystretch}{1.0}
\caption{The Prompt for Rollout}
\label{tab:rollout_prompt}
\begin{tabular}{p{\linewidth}}
\toprule
\textbf{System} \\
You are an expert in mobile phone operations and UI automation.
Your task is to analyze a mobile operation trajectory and summarize what worked and what did not,
focusing on concise and actionable insights. \\
\midrule

\textbf{User} \\

\textbf{Task Info}: 
App: \{app\};
Intent: \{intent\_category\};
Instruction: \{instruction\}. \\

\textbf{Mobile Operation Trajectory}: A sequence of steps, each containing
\textit{Thought} (planned intent),
\textit{Action} (Tap, Swipe, Type, etc.),
and \textit{Summary} (on-screen result or observation). \\

\{trajectory\} \\

\textbf{Your Task}: Analyze the trajectory and provide a concise summary (\textbf{under 200 words}) covering: \\

\textbf{(1) Successful Actions}: UI interactions that worked as intended; \\
\textbf{(2) Failed Actions}: interactions that did not achieve the goal; \\
\textbf{(3) UI Understanding}: correctness of UI element identification and interaction; \\
\textbf{(4) Navigation Pattern}: whether the navigation sequence was reasonable; \\
\textbf{(5) User Patterns}: recurring app preferences or frequent behaviors; \\
\textbf{(6) Anti-Patterns}: actions leading to loops or failures, repeated ineffective taps, or cases where continued interaction was counterproductive. \\

\textbf{Output Format}: Step-by-step summary: \\
Step 1: [Action and outcome] \\
Step 2: [Action and outcome] \\
\ldots \\
Overall: [Key insights explaining what worked or failed and why]. \\
\bottomrule
\end{tabular}
\end{table*}

\begin{table*}[t]
\centering
\setlength{\tabcolsep}{4pt}
\renewcommand{\arraystretch}{1.0}
\caption{The Prompt for Single Query Critique}
\label{tab:single_query_critique_prompt}
\begin{tabular}{p{\linewidth}}
\toprule
\textbf{System} \\
You are an expert in mobile UI automation and user behavior analysis.
Your task is to compare multiple attempts at the same mobile task and extract actionable lessons,
with special focus on learning user preferences and habits.\\
\textbf{CRITICAL}: Output MUST be wrapped in \texttt{<Experiences></Experiences>} tags. \\
\midrule

\textbf{User} \\

\textbf{Task Info}: 
App: \{app\};
Intent: \{intent\_category\};
Instruction: \{instruction\}. \\

\textbf{Multiple Attempts}: Summaries of repeated executions:
\{attempts\}. \\

\textbf{Your Task}: Compare attempts and extract \textbf{5--8 actionable experiences} for future tasks, focusing on: \\

\textbf{(1) User Preferences} (priority): preferred apps, frequent destinations, shopping/music habits, routines, inferred percentages; \\
\textbf{(2) UI Navigation}: locations of key UI elements and optimal paths; \\
\textbf{(3) Action Sequences}: correct operation order, timing, recovery; \\
\textbf{(4) Element Identification}: distinguishing similar or dynamic UI elements; \\
\textbf{(5) Context Awareness}: app-specific behaviors, success/failure cases; \\
\textbf{(6) Task Completion}: screens/states indicating completion and when to STOP. \\

\textbf{Output Format (MANDATORY)}: \\

\texttt{<Experiences>} \\
1. [User Preference -- \{intent\_category\}] <specific preference (percentages if known)> \\
2. [UI Navigation -- \{app\}] <key UI element locations> \\
3. [Action Sequence -- \{intent\_category\}] <operation order> \\
4. [Element Identification -- \{app\}] <element identification guidance> \\
5. [Context] <when approaches work or fail> \\
6. [Task Completion -- \{intent\_category\}] <completion signals> \\
\texttt{</Experiences>} \\

\textbf{Reminders}: start with user preferences; one experience per line; be specific and actionable; include percentages when inferable. \\
\bottomrule
\end{tabular}
\end{table*}

\begin{table*}[t]
\centering
\setlength{\tabcolsep}{4pt}
\renewcommand{\arraystretch}{1.0}
\caption{The Prompt for Group Experience Update}
\label{tab:group_experience_update_prompt}
\begin{tabular}{p{\linewidth}}
\toprule

\textbf{System} \\
You are an expert in knowledge management for mobile automation.
Your task is to process a batch of new experiences from a single task and decide how to integrate them with existing experiences.
\textbf{CRITICAL}: Each line in the new experiences is ONE independent experience and must be processed separately.
You must output valid JSON only, with no additional text or explanations. \\

\midrule

\textbf{User} \\

\textbf{Existing Experiences Database}: \\
\{existing\_experiences\} \\

\textbf{New Experiences}: Newly extracted experiences from recent task attempts.
Each numbered line represents ONE independent experience. \\
\{new\_experiences\} \\

\textbf{Your Task}: For EACH new experience, choose exactly one operation: \\

\textbf{ADD}: completely new information not covered by existing experiences; \\
\textbf{UPDATE}: refines, corrects, or expands an existing experience (specify the ID to replace); \\
\textbf{DISCARD}: redundant with existing experiences or not useful. \\

\textbf{Important Rules}: \\
- User Preference experiences: \textbf{ALWAYS ADD or UPDATE}; \\
- Similar experiences with new information (e.g., new items or updated percentages): use UPDATE; \\
- Completely redundant experiences: use DISCARD; \\
- Each operation must reference exactly ONE new experience line. \\

\textbf{Output Format (JSON ONLY)}: Return ONLY valid JSON in the following structure: \\

\texttt{[} \\
\quad \texttt{\{ "operation": "ADD", "content": "[User Preference -- Music] User consistently uses QQ Music for music playback", "reason": "New preference" \},} \\
\quad \texttt{\{ "operation": "UPDATE", "id": "G3", "content": "[UI Navigation -- QQ Music] Search icon at top-right; tap to activate keyboard", "reason": "Refines existing experience" \},} \\
\quad \texttt{\{ "operation": "DISCARD", "content": "[Context] Wait for page to load", "reason": "Redundant information" \}} \\
\texttt{]} \\

\textbf{CRITICAL}: Output must be valid JSON parsable by \texttt{json.loads()}, with no text before or after the array. \\
\bottomrule
\end{tabular}
\end{table*}

\begin{table*}[t]
\centering
\setlength{\tabcolsep}{4pt}
\renewcommand{\arraystretch}{1.0}
\caption{The Prompt for Batch Experience Update}
\label{tab:batch_experience_update_prompt}
\begin{tabular}{p{\linewidth}}
\toprule

\textbf{System} \\
You are an expert in knowledge management for mobile automation.
Your task is to create a comprehensive, non-redundant experience base that captures both operational knowledge and user preferences.
You must output valid JSON only, with no additional text or explanations. \\

\midrule

\textbf{User} \\

\textbf{Current Experiences and Proposed Operations}: \\
\{experiences\_and\_operations\} \\

\textbf{Your Task}: Review all proposed operations and produce a final revision plan that: \\
(1) resolves conflicts between operations;
(2) removes redundancies by merging similar experiences;
(3) ensures experiences are specific and actionable;
(4) maintains a clean and organized experience base;
(5) prioritizes user preference experiences, which should always be preserved and updated. \\

\textbf{Guidelines for High-Quality Experiences}: \\
- Keep each experience concise (1--2 sentences, max 50 words); \\
- Each experience should be independently useful; \\
- Prefer specific, actionable guidance over generic advice; \\
- Use consistent prefixes: [Category] or [Category -- SubCategory]; \\
- User Preference experiences MUST use format: [User Preference -- IntentCategory]. \\

\textbf{Output Format (JSON ONLY)}: Return ONLY valid JSON in the following structure: \\

\texttt{[} \\
\quad \texttt{\{ "operation": "ADD", "content": "New experience with proper [Category] prefix" \},} \\
\quad \texttt{\{ "operation": "UPDATE", "id": "G2", "content": "Updated experience replacing G2" \},} \\
\quad \texttt{\{ "operation": "DELETE", "id": "G5" \}} \\
\texttt{]} \\

\textbf{Operation Types}: \\
\textbf{ADD}: create a new experience;
\textbf{UPDATE}: replace an existing experience by ID;
\textbf{DELETE}: remove an existing experience by ID. \\

\textbf{CRITICAL RULES}: \\
- Output must be valid JSON parsable by \texttt{json.loads()}; \\
- Do not include any text before or after the JSON array; \\
- All content fields must be complete experience strings with [Category] prefixes; \\
- IDs must exactly match existing experience IDs (e.g., G2, G15); \\
- Never delete user preference experiences unless absolutely necessary. \\
\bottomrule
\end{tabular}
\end{table*}

\begin{table*}[t]
\centering
\setlength{\tabcolsep}{4pt}
\renewcommand{\arraystretch}{1.0}
\caption{The Prompt for Problem Solving with Learned Experiences}
\label{tab:problem_with_experience_prompt}
\begin{tabular}{p{\linewidth}}
\toprule

\textbf{Learned Experiences} \\
\{experiences\} \\

\textbf{Your Task} \\
\{problem\} \\

\textbf{Important Instructions} \\
Use the learned experiences above to guide your operations. They contain valuable insights from previous successful attempts and documented user preferences. \\

Pay special attention to User Preference experiences- they tell you which apps, locations, or items the user commonly prefers. When multiple options are available, ALWAYS choose based on documented user preferences.
\textbf{Example}:\\
If experiences indicate ``User prefers QQ Music (75\%)'', you should use QQ Music for music-related tasks by default. \\

\bottomrule
\end{tabular}
\end{table*}

\section{Examples of Learned Experiences}
In this appendix, we provide some examples in table \ref{tab:experience_en_id_single}, which are extracted from 49 learned experiences by Me-Agent.
\begin{table*}[t]
\centering
\caption{Sample Experiences Extracted from User Interaction Trajectories (ID included)}
\label{tab:experience_en_id_single}
\begin{tabular}{p{16cm}}
\toprule
\textbf{Experience (ID + English Translation)} \\
\midrule

G0: [User Preference - Movies] User prefers accessing movie and TV series content through Douban's ``Books, Movies, and TV'' section for ratings and rankings. \\

G1: [UI Navigation - Douban] Navigate to ``Books, Movies, and TV'', then ``Douban Rankings'', and switch to ``Movie Rankings'' or ``TV Series Rankings'' for rankings. \\

G3: [Error Avoidance] Use ``Tap\_Type\_and\_Enter'' to efficiently input text and avoid repeated tapping; ensure the search bar is engaged before typing or using shortcuts. \\

G5: [UI Navigation - Xiaohongshu] The search icon is located at the top-right corner, serving as a reliable starting point for searches. \\
G8: [User Preference - Shopping] User prefers the JD app for precise product searches such as MacBook Air. \\

G10: [Action Sequence - JD] Open the JD app → Tap the search bar → Clear existing text → Type ``latest iPad'' → Execute the search. \\

G11: [Task Completion - JD] The task is complete when the search results display the latest iPad models. \\

G23: [User Preference - Music] User prefers QQ Music for music-related tasks, indicating a strong preference for this app. \\

G24: [Action Sequence - Music] Open QQ Music → Tap the search bar → Enter the song name or select from search history → Tap the play button next to the song. \\

G25: [Task Completion - Music] The task is complete when the song begins playing. \\

G34: [User Preference - Reading] User prefers using the WeChat Reading app for novels, indicating a strong preference for this application. \\
G42: [User Preference - Posts] User frequently searches for ``weight loss tips'' in Xiaohongshu, indicating a strong interest in this content. \\

G43: [Action Sequence - Posts] Open Xiaohongshu → Tap the search icon → Utilize search history by selecting ``weight loss tips''. \\

G44: [Context] Utilizing search history is effective when the user has previously searched for the same term. \\

G45: [Task Completion - Posts] The task is complete when the search results for ``weight loss tips'' are displayed. \\

\bottomrule
\end{tabular}
\end{table*}

\section{ Examples of Dataset}

In this appendix, we provide an example user from User FingerTip, including both training and testing sets. 
Table \ref{tab:user5} shows the training set tasks for a randomly selected user, with each task specifying the instruction, intent category, and target app. 
Table \ref{tab:user5_evaluate_combined_gt} presents the corresponding testing set tasks, where each test task is divided into Type~I (APP-Ambiguous Instruction) and Type~II(Content-Ambiguous Instruction), along with the target app and the ground truth content.

\begin{table*}[htbp]
    \centering
    \small
\caption{Training set instructions for an example user}
    \label{tab:user5}
    \begin{tabular}{
        >{\arraybackslash}m{3.5em}    
        >{\arraybackslash}m{25em}     
        >{\arraybackslash}m{5em}      
        >{\arraybackslash}m{5em}}     
        \toprule
        \textbf{Task ID} & 
        \textbf{Instruction} & 
        \textbf{Intent Category} & 
        \textbf{Target App} \\
        \midrule
        train\_0 & Open WeChat Reading to read The Three-Body Problem & Reading & WeChat Reading \\
        \midrule
        train\_1 & Open WeChat Reading to read Xiangtu Zhongguo & Reading & WeChat Reading \\
        \midrule
        train\_2 & Open Xiaohongshu to search posts about home bartending & Posts & Xiaohongshu \\
        \midrule
        train\_3 & Open Xiaohongshu to search travel guides for Summer Palace & Posts & Xiaohongshu \\
        \midrule
        train\_4 & Open Xiaohongshu to search fitness strategies and open the first one & Posts & Xiaohongshu \\
        \midrule
        train\_5 & Open JD.com to check price of latest iPad & Shopping & JD.com \\
        \midrule
        train\_6 & Open JD.com to search for price of monitors & Shopping & JD.com \\
        \midrule
        train\_7 & Open JD.com to search for price of latest MacBook Air & Shopping & JD.com \\
        \midrule
        train\_8 & Open Douban to check currently high-rated movies & Posts & Douban \\
        \midrule
        train\_9 & Open QQ Music to play the song "My Future" & Music & QQ Music \\
        \midrule
        train\_10 & Open QQ Music to play the song "Replicate Memories" & Music & QQ Music \\
        \midrule
        train\_11 & Open QQ Music to play the song "The Best Day" & Music & QQ Music \\
        \midrule
        train\_12 & Open Douban to check the currently highest-rated TV dramas and their reviews & Posts & Douban \\
        \midrule
        train\_13 & Open Xiaohongshu to search for body fat rate related to fitness & Posts & Xiaohongshu \\
        \midrule
        train\_14 & Open Xiaohongshu to search for fat loss tips & Posts & Xiaohongshu \\
        \bottomrule
    \end{tabular}
\end{table*}

\begin{table*}[htbp]
    \centering
    \small
    \caption{Testing set instructions for an example user.}
    \label{tab:user5_evaluate_combined_gt}
    \begin{tabular}{
        >{\arraybackslash}m{3.5em}   
        >{\arraybackslash}m{6em}      
        >{\arraybackslash}m{11em}     
        >{\arraybackslash}m{13em}     
        >{\arraybackslash}m{5em}      
        >{\arraybackslash}m{10.5em}}  
        \toprule
        \textbf{Task ID} & 
        \textbf{Intent Category} &
        \textbf{Type~I Instruction} & 
        \textbf{Type~II Instruction} & 
        \textbf{Target App} &
        \textbf{Target Content} \\
        \midrule
        test\_1 & Posts   & Search for posts related to fitness strategies & Search for sports strategies that interest me & Xiaohongshu & Sports strategies the user likes: fitness-related activities \\
        \midrule
        test\_2 & Shopping & Check the price of iPhone 16 Pro Max & Check the prices of digital products that interest me & JD.com & Digital products the user likes: iPad, monitor, Mac \\
        \midrule
        test\_3 & Music    & Play the song "Chun Ni (Spring Mud)" & Play music that I like & QQ Music & Music the user likes: My Future, Replicate Memories \\
        \midrule
        test\_4 & Posts    & Check the highest-rated movies of 2024 and their reviews & Check the movies I follow and their reviews & Douban & No restrictions, just open what the user follows \\
        \midrule
        test\_5 & Reading  & Read "Xiangtu Zhongguo" & Read books that I like & WeChat Reading & Books the user likes: The Three-Body Problem, Xiangtu Zhongguo \\
        \bottomrule
    \end{tabular}
\end{table*}

\end{document}